\documentclass[conference]{IEEEtran}
\IEEEoverridecommandlockouts

\usepackage{nicefrac}
\usepackage{caption}
\usepackage{cite}
\usepackage{amsmath,amssymb,amsfonts,algcompatible}
\usepackage{graphicx}
\usepackage{textcomp}
\usepackage{xcolor}
\usepackage{soul}
\usepackage{xspace}
\usepackage{subfigure}
\usepackage{algpseudocode}

\usepackage[ruled,linesnumbered]{algorithm2e}
\SetKwComment{Comment}{$\triangleright$\ }{}
\newlength\mylenin
\newcommand\myinput[1]{%
\settowidth\mylenin{\KwIn{}}%
\setlength\hangindent{\mylenin}%
\hspace*{\mylenin}#1\\}

\let\oldnl\nl 
\newcommand{\nonl}{\renewcommand{\nl}{\let\nl\oldnl}} 

\newlength\mylenout

\usepackage{booktabs}
\usepackage{threeparttable}
\def\BibTeX{{\rm B\kern-.05em{\sc i\kern-.025em b}\kern-.08em
    T\kern-.1667em\lower.7ex\hbox{E}\kern-.125emX}}

\definecolor{stelios_colour}{RGB}{200, 238, 200}
\definecolor{light_red}{RGB}{255, 204, 204}

\definecolor{crimson}{rgb}{0.86, 0.08, 0.24}

\newcommand{\tool}{MultiTASC++\xspace}

\newif\ifcomment

\commenttrue

\ifcomment
\newcommand{\stelios}[1]{\sethlcolor{stelios_colour}\hl{[\textbf{Stelios:} #1]}}
\newcommand{\sokratis}[1]{\sethlcolor{orange}\hl{[Sokratis: #1]}}

\else
\newcommand{\stelios}[1]{}
\newcommand{\sokratis}[1]{}

\fi

\begin{document}

\title{\tool: A Continuously Adaptive Scheduler for Edge-Based Multi-Device Cascade Inference}

\author{Sokratis~Nikolaidis\IEEEauthorrefmark{2},
        Stylianos~I.~Venieris\IEEEauthorrefmark{3},
        and~Iakovos~S.~Venieris\IEEEauthorrefmark{2}%
\\

\IEEEauthorblockA{\IEEEauthorrefmark{2}National Technical University of Athens, Athens, Greece, \IEEEauthorrefmark{3}Samsung AI Center, Cambridge, UK
}
\IEEEauthorblockA{Email: sokratisnikolaidis@mail.ntua.gr, s.venieris@samsung.com, venieris@cs.ece.ntua.gr}
\vspace{-0.8cm}
}

\maketitle

\begin{abstract}
    Cascade systems, consisting of a lightweight model processing all samples and a heavier, high-accuracy model refining challenging samples, have become a widely-adopted distributed inference approach to achieving high accuracy and maintaining a low computational burden for mobile and IoT devices. As intelligent indoor environments, like smart homes, continue to expand, a new scenario emerges, the multi-device cascade. In this setting, multiple diverse devices simultaneously utilize a shared heavy model hosted on a server, often situated within or close to the consumer environment. This work introduces \tool, a continuously adaptive multi-tenancy-aware scheduler that dynamically controls the forwarding decision functions of devices to optimize system throughput while maintaining high accuracy and low latency. Through extensive experimentation in diverse device environments and with varying server-side models, we demonstrate the scheduler's efficacy in consistently maintaining a targeted satisfaction rate while providing the highest available accuracy across different device tiers and workloads of up to 100 devices. This demonstrates its scalability and efficiency in addressing the unique challenges of collaborative DNN inference in dynamic and diverse IoT environments.
\\
\end{abstract}


\vspace{-0.4cm}
\section{Introduction}
\label{sec:intro}
In the last few years, there have been notable advancements in the realm of on-device execution for Deep Learning (DL) inference tasks~\cite{oodin2021smartcomp}. Concurrently, the proliferation of indoor intelligent environments~\cite{laskaridis2022future}, encompassing smart homes and offices, presents an opportunity for DL to facilitate novel applications across a diverse range of smart devices like IoT cameras and AI speakers. However, due to their compact form-factor and energy-consumption constraints, the majority of these devices fall within the low-end of the computational spectrum. Unlike high-end smartphones equipped with robust processors and accelerators such as GPUs and NPUs~\cite{smart2021imc}, low-end devices lack the capability to deploy state-of-the-art Deep Neural Networks (DNNs). Consequently, they resort to lightweight models, albeit with lower accuracy.

Considering the drawbacks associated with offloading data to the cloud for inference, such as increased bandwidth usage, latency, and privacy concerns, an alternative strategy has been gaining ground. This new strategy involves placing the server within or in close proximity to the consumer environment, often in the form of a dedicated AI hub designed to assist nearby devices~\cite{laskaridis2022future}. Within this framework, cascade architectures have emerged as a notable deployment approach~\cite{park2015big,li2021appealnet,wang2017idk,mirzadeh2020optimal,stamoulis2018designing,kouris2018cascade}. These architectures capitalize on the inherent variability in sample difficulty, opting to process only the more challenging cases with a robust server-based model, while delegating the processing of simpler samples, which typically constitute the majority of the data stream, to on-device execution using a lightweight model. A substantial body of research has delved into cascade architectures, with a primary focus on refining the forwarding decision criteria and optimizing the selection of model pairs to enhance overall efficacy.

Despite the advancements made, the prevailing focus of existing work has been confined to scenarios where a \textit{single} device utilizes the server at any given moment. This assumption is no longer applicable in the context of emerging intelligent environments, where multiple devices concurrently undertake DL inference tasks with the support of a shared AI hub~\cite{NaShFuKaSh2022}. This scenario introduces the novel setting of \textit{multi-device cascade}, wherein multiple devices utilize the same model on a shared edge-based server. A system operating in this mode must exhibit scalability with regard to the number of devices, effectively balancing rapid response times and high accuracy across the devices. Conventional approaches, which treat each model cascade independently, would either lead to brute-forcing inference requests through the server's resources, resulting in system overload, or force all devices to resort to on-device execution, nullifying any accuracy advantages. Consequently, there is a pressing need for innovative methods explicitly tailored to address the challenges posed by a multi-device cascade.

In this setting, the current state-of-the-art work, MultiTASC~\cite{multitasc2023iscc}, introduced a multi-tenancy-aware scheduling approach for multi-device cascades. MultiTASC proposed a scheme where forwarding decision functions can be dynamically reconfigured at runtime, providing the adaptability missing from prior cascade architectures. The decision to modify the forwarding functions would be based on the monitoring of the server-hosted model's batch size, regarded as a metric for the server's load. As such, the scheduler would tune the decision functions when the running batch size deviated from a predefined optimal value that was calculated during initialization. 

Despite its improvements over conventional cascade systems, MultiTASC resulted in an overly relaxed policy before a certain number of devices was reached and an excessively strict one after the influx of requests became considerably high. Furthermore, when run across multiple independent runs, the observed system behavior demonstrated noticeable variance, indicating that a more fine-grained monitoring approach was needed to ensure a robust deployment.

In this work, we propose \tool, a continuously adapting multi-tenancy-aware scheduler specifically designed to address the challenges inherent in deploying multi-device cascade architecture in high-demand, AI-enabled indoor spaces. Building upon the strengths of MultiTASC, we retain the dynamically reconfigurable forwarding functions across client devices, aiming to control the server's inference request rate at runtime. Departing from MultiTASC, we introduce a new method for tuning the decision functions that allows for more fine-grained and continuous adaptation. 
The key contributions of this paper are the following:
\begin{itemize}
    \item A system model of the multi-device cascade architecture. By expanding the cascade architecture to accommodate multiple devices, our parametrization exposes the tunable parameters and enables system designers to systematically investigate its trade-offs.
    \item A new multi-tenancy-aware scheduler optimized for the multi-device cascade architecture. With its enhanced approach of reconfiguring the forwarding decision functions, we consider each device's latency requirements independently leading to more effective, device-tailored adaptation. We further introduce the continuous, rather than in discrete steps, tuning of the decision functions, resulting in finer-grained adaptability. Lastly, we introduce server model switching, where the server-side model can be dynamically swapped for another with a different latency-accuracy trade-off. In this manner, we add a new design dimension in the multi-device cascade architecture, further increasing its adaptability.
\end{itemize}
In the following section, we present the current state of AI-focused edge computing and discuss related work. In Section~\ref{sec:3_methodology}, we describe the system architecture in the multi-device cascade setting as well as our formulation of the target problem. Section~\ref{sec:4_scheduler} presents the novel \tool scheduler and its internal design, followed by the experimental evaluation in Section~\ref{sec:5_evaluation}. We conclude with Section~\ref{sec:7_conclusion}, where we summarize the proposed approach and outline possible avenues for future research.

\section{Background \& Related Work}
\label{sec:related_work}

\begin{figure*}[t]
\includegraphics[width=0.75\textwidth]{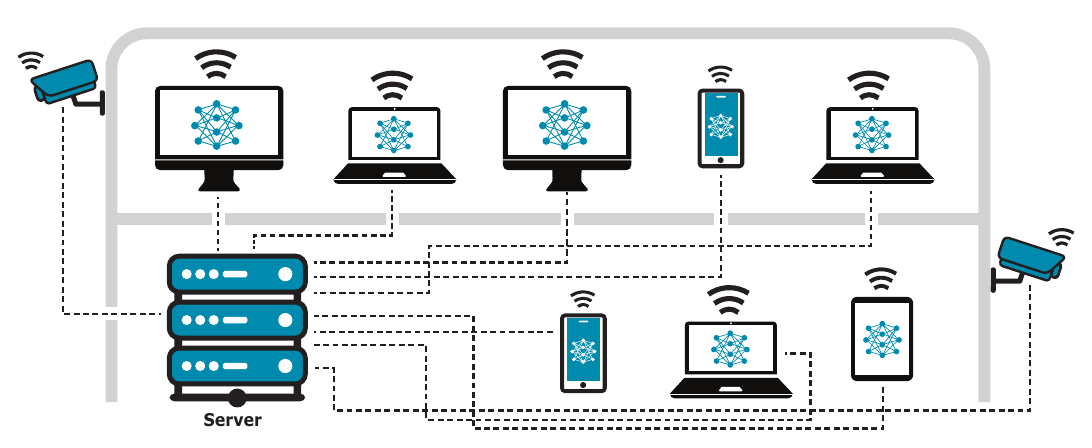}
\centering
\caption{Example of an AI-driven smart office.}
\label{fig:smartoffice}
\end{figure*}

\subsection{On-device DNN inference}
In recent years, there has been an explosion of Artificial Intelligence (AI) applications and services thanks to significant advancements in DL. These applications span a wide range, from personal assistants and recommendation systems to autonomous vehicles and healthcare diagnostics. Furthermore, the widespread adoption of mobile computing and the Internet of Things (IoT) has led to billions of interconnected mobile and IoT devices, collectively generating an immense volume of data at the network edge \cite{ZhiEdgeIntelligence2019}. This has created the need to push the execution of DNN applications at the edge of the network leading to a substantial surge in the deployment of DL models on resource-constrained devices \cite{oodin2021smartcomp}. Although on-device training of DL models remains a challenging endeavor due to the limited computational resources and memory constraints of such devices, significant progress has been made in on-device inference. Several techniques have been proposed to enable efficient on-device inference, including:

\textbf{Lightweight model design:} One approach to enable on-device inference is the design of lightweight DL models. Models like MobileNetV2 \cite{sandler2018mobilenetv2}, EfficientNet-Lite \cite{tan2019effnet} and NasNet-Mobile \cite{zoph2017LearningTA} have been specifically crafted to achieve high accuracy while minimizing computational and memory requirements.

\textbf{Model quantization:} Quantization techniques \cite{han2015deep} reduce the precision of model weights and activations, effectively decreasing memory and computational demands without substantial loss in accuracy.

\textbf{Model pruning:} Pruning methods, such as channel pruning \cite{he2017channel}, aim to reduce the size of DNNs by removing unimportant neurons, thereby reducing computational overhead.

\textbf{Knowledge distillation:} Knowledge distillation \cite{hinton2015distilling} involves training a smaller, more efficient model to mimic the predictions of a larger, complex model. This allows for the transfer of knowledge from larger models to smaller ones.

\textbf{Optimized scheduling:} Scheduling and runtime optimization techniques \cite{CoBiCh2022} help allocate computational resources efficiently, ensuring that DL inference tasks run smoothly on constrained devices.

Despite these advancements, modern intelligent environments like smart homes and offices are often equipped with small-form-factor, resource-constrained devices (e.g., smart cameras, AI speakers). These devices lack the processing power to support high-accuracy, computationally-intensive models, driving the need for distributed collaborative inference approaches.

\subsection{Distributed collaborative inference}
Distributed collaborative inference systems leverage a central server to assist mobile and embedded devices in performing DL inference tasks. Notably, the server can be strategically placed at the network edge, close to the devices, to minimize latency and optimize real-time processing. Two primary schemes have emerged in this domain: offloading and cascading.
\subsubsection*{A. Offloading}
Offloading techniques aim to distribute the computational load between the device and the server. In this scheme, the DNN is divided into two parts, with the initial part executed on the device and the latter part on the server. One standout contribution in this domain is the Neurosurgeon \cite{kang2017neurosurgeon} framework, which focuses on the selection of a singular split point for offloading Convolutional Neural Networks (CNNs) from devices to servers, with the objective of optimizing either latency or energy consumption. Later work such as \cite{li2019jalad, almeida2022dyno} explore the trade-off between latency and accuracy that is introduced when taking the offloading decision. \cite{laskaridis2020spinn, huang2020clio} try to address the offloading dilemma in a progressive manner, requiring some training before deployment. Offloading approaches have managed to alleviate some of the computational burdens on the device while maintaining the accuracy of a complex model.
\subsubsection*{B. Cascades} Cascade schemes involve a sequence of DNNs with progressively increasing complexity and accuracy. After processing input data through a model, a forwarding decision function determines whether to continue with the current result or proceed to the next, more complex model. \cite{park2015big} is one of the first notable contributions on this approach. It introduces a static technique, calculating an optimized threshold before runtime, as well as a dynamic technique, aiming to optimize the threshold at runtime. The forwarding function in both techniques is defined by analyzing the differences between the softmax results. From our experiments, \cite{park2015big} leads to polarized execution in the majority of cases, with the data being processed either solely locally by the light model or on the server side by the heavy model, \textit{i.e.}~without a balanced split between device and server. \cite{li2021appealnet} proposed a trainable forwarding criterion based on a neural head attached to the light model's feature extractor. This approach yields great results when tested on a certain pair of CNNs but requires training before deployment for every combination of networks.  \cite{wang2017idk} investigates the use of multiple DNN models going beyond two element cascades, coupled with the evaluation of various decision metrics. Additionally, solutions have been proposed to deploy cascades under tight energy constraints \cite{mirzadeh2020optimal, stamoulis2018designing}. Most of the aforementioned work focus on the image classification task and don't consider the problem of dynamic adaptation at runtime. The task of video classification has also been studied \cite{kang2017noscope, shen2017FastVC} and lately, the first attempts to accommodate Large Language Models have emerged \cite{wang2023tabi}. 
\subsection{Multi-device cascades}
Previous research on cascade architectures is predominantly centered around isolated scenarios, where a single device enjoys exclusive access to a dedicated server. Nevertheless, this assumption no longer aligns with the reality of AI-driven indoor environments like the one in Fig.~\ref{fig:smartoffice}. The pervasive integration of AI technologies has given rise to an expanding array of AI-powered devices, resulting in a pressing demand for simultaneous support from a shared server. 

In these complex and interconnected scenarios, a more nuanced examination is indispensable. Hasty or simplistic deployment strategies risk overloading the server and causing protracted response times. However, relying solely on local execution can severely compromise the overall accuracy of the system, as it misses out on the collaborative potential of cascaded processing. Thus, it becomes evident that as we navigate the intricacies of AI-driven indoor environments, thoughtful strategies are paramount to harnessing the full potential of these technologies, optimizing both efficiency and accuracy.

This paper addresses the unexplored setting of multi-device cascades, where multiple devices operate simultaneously, sharing an edge server assisting in the execution of DL inference tasks. This paper provides a principled approach to tackling the challenges of resource allocation and model selection in this complex scenario, facilitating the straightforward and adaptable deployment of such an architecture.

\section{Multi-Device Cascade of Classifiers}
\label{sec:3_methodology}
\begin{figure}[t]
\includegraphics[width=\columnwidth]{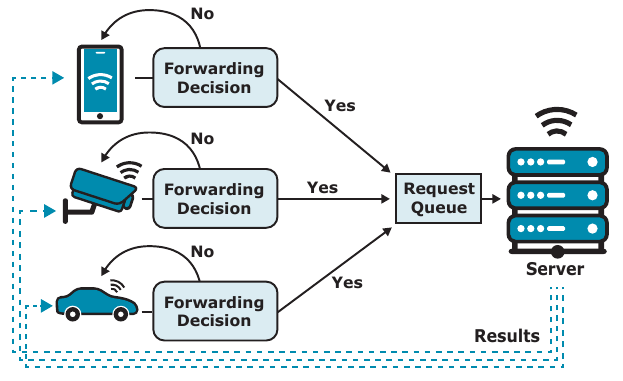}
\centering
\caption{System architecture of a multi-device cascade~\cite{multitasc2023iscc}.}
\label{fig:multicasc}
\end{figure}

In Fig.~\ref{fig:multicasc}, we present the comprehensive system architecture of a multi-device cascade, specifically designed for executing DL inference tasks on IoT devices in a collaborative setting. Within this architecture, all IoT devices are engaged in performing a common task, such as object detection, albeit they may host different DL models tailored to their computational capabilities and requirements.
The main components of this system architecture include the following.

\textbf{IoT devices:} These devices are the primary endpoints where the DL inference tasks are executed. Each IoT device is equipped with its own DL model designed to process incoming data efficiently. These devices generate predictions independently based on their respective models.

\textbf{Device-hosted models:} DL models on the devices are trained on the same task, operating independently in all other aspects. Each model can be of different architecture, complexity and efficiency, depending on the computational resources and demands of the respective device. Thus, it is of paramount importance that the scheduler maintains high performance in scenarios of device heterogeneity. 

\textbf{Forwarding decision function:} The output predictions generated by each IoT device are first given to a forwarding decision function. This decision function assesses the confidence of the DL model's output on each device. If the model is sufficiently confident in its prediction, the result remains unchanged, and no further action is taken. However, if there is uncertainty or low confidence in the prediction, the sample is earmarked for further analysis.

\textbf{Server:} Samples that require additional scrutiny are forwarded to a centralized server for in-depth processing. The server hosts a more accurate and computationally-intensive DL model capable of refining the predictions made by the IoT devices.

\textbf{Request queue:} The forwarded samples from all IoT devices are temporarily stored in a request queue at the server. This queue serves as a staging area where samples awaiting processing are collected. The request queue ensures efficient and organized data flow from the IoT devices to the server.

\textbf{Server-hosted model:} The server-side model processes the samples drawn from the request queue. This model is shared among all connected IoT devices, allowing for collaborative refinement of predictions by leveraging the advanced "knowledge" of the network.

\textbf{Result distribution:} Finally, the results produced by the server-side model are distributed back to their corresponding IoT devices as soon as they become available. This seamless distribution of refined predictions ensures that each IoT device benefits from the improved accuracy achieved by the server.
\begin{figure*}[t!]
\includegraphics[width=0.75\textwidth]{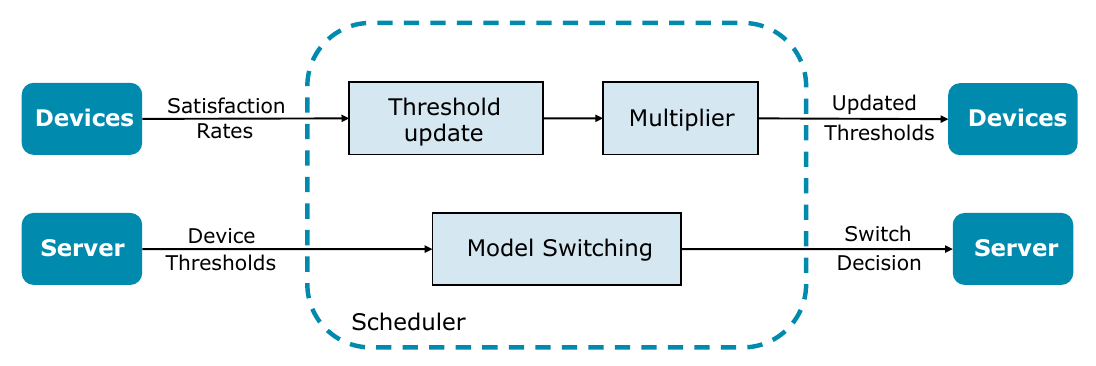}
\centering
\caption{Architecture of the \tool scheduler.}
\label{fig:scheduler}
\end{figure*}
\subsection{Single-device cascade}
Let us consider a single IoT device running a classification-based DL inference task. Let $x$$\in$$\mathcal{X}$ be the input and $y$$\in$$\{1, ..., K \}$ the classification label produced by the model, where $K$ is the number of classes. The confidence of the given DL model on its output $y$ can be calculated by using the decision function $d(\cdot)$. The output of $d(\cdot)$ is binary; for $d(\cdot)=0$ we conclude that the model is confident and therefore the result is acceptable, whereas for $d(\cdot)=1$ the sample $x$ should be forwarded to the server for further processing. Denoting the classification function of the light model by \mbox{$f_l: \mathcal{X} \rightarrow [0,1]^{K}$} that yields the softmax output vector of the model whose maximum value is the predicted class, and the classification function of the heavy model by \mbox{$f_h: \mathcal{X} \rightarrow [0,1]^K$}, we formally define a collaborative cascade system as: 
\begin{equation*}
\it{casc}_{f_l, f_h, d} (x) =  
 \begin{cases} 
      f_l(x) &\text{if} \quad \, d(f_l(x)) = 0 \\
      f_h(x) &\text{if} \quad \, d(f_l(x)) = 1 
\end{cases}
\label{cascade}
\end{equation*}
\subsection{Multi-device cascade} To capture multi-device cascade architectures (Fig.~\ref{fig:multicasc}), we expand the existing single-device cascade system representation as follows. Let $\mathcal{D}$ be the set of devices assisted by the same server. Then, the multi-device cascade system is defined as:
\begin{gather*}
    \small
    \it{casc}_{f_l^i, f_h, d^i} (x^i) =  
    \begin{cases} 
          f_l^i(x^i) &\text{if} \quad  \, d^i(f_l^i(x^i)) = 0 \\
          f_h(x^i) &\text{if}  \quad \, d^i(f_l^i(x^i)) = 1 
    \end{cases}
    \\ \forall i \in \{1, ..., |\mathcal{D}|\}
    \label{multicascade}
\end{gather*}
where $x^i$$\in$$\mathcal{X}^i$ is a sample processed by the i-th device, $f_l^i$ the classification function of the DL model deployed on the i-th device, $f_h$ the shared heavy model on the server, and $d^i(f_l^i(x^i))$ the forwarding decision function of the i-th device. Note that the shared heavy model $f_h$, is the only variable not depending on the devices.
\noindent
\subsection{Congestion problem}
In the case of the single-device cascade, the server's computational resources are exclusively assisting a single device, resulting in minimal response time and preservation of accuracy. However, in the area of IoT spaces it is increasingly important to leverage a server's capabilities across multiple devices simultaneously, hence implementing a multi-device cascade. This approach paves the way towards amortizing the server's cost and maximizing its utility. Nonetheless, depending on the specific conditions, if the arrival rate of incoming requests exceeds the server's processing throughput capacity, the server becomes overwhelmed, leading to extended waiting times for requests in the queue.

For a given number of devices denoted by $|\mathcal{D}|$, Eq.~\ref{serverAR}~expresses the arrival rate of requests to the server, \textit{i.e.}~the rate at which results are deemed as unsatisfactory by the decision functions on the devices and are therefore forwarded to the server.
\begin{equation}
    AR_{\text{server}} = \sum_{i~=~1}^{|\mathcal{D}|} \frac{p_{\text{casc}}^i}{t_{\text{inf}}^i}
    \label{serverAR}
\end{equation}
where $t_{\text{inf}}^i$ is the average inference latency of a sample on the i-th device and $p_{\text{casc}}^i$ is the probability of a sample giving $d^i(f_l^i(x^i))=1$. 

Given the attainable throughput $T_{\text{server}}$ of the server, we distinguish three different states:
\begin{itemize}
    \item $AR_{\text{server}} < T_{\text{server}}$: The processing rate of the server exceeds the arrival rate, leading to the server being underutilized. Forwarding a greater quantity of challenging samples to the server could enhance accuracy.
    \item $AR_{\text{server}} = T_{\text{server}}$: A state of equilibrium is reached, where requests are promptly processed upon arrival, preventing accumulation, and ensuring full utilization of the server's processing power. 
    \item $AR_{\text{server}} > T_{\text{server}}$: The rate of incoming requests surpasses the server's processing capacity. If this condition persists, it will result in a substantial accumulation of requests in the queue, leading to excessive latency.
\end{itemize}
The $p_{\text{casc}}^i$ of the forwarding decision function is not static since it depends on the processing order of the samples. Therefore, this architecture involves stochastic components at runtime and could greatly benefit by dynamically adapting its state depending on the current conditions. Since $t_{\text{inf}}^i$ and $T_{\text{server}}$ are fixed based on the device and server-side processors, we opt to manipulate $p_{\text{casc}}^i$ by changing the parameters of $d^i(f_l^i(x^i))$ in order to introduce adaptability to the system.
\noindent
\subsection{Problem optimization} We frame the aforementioned setting as a multi-objective optimization problem, seeking to maximize accuracy and throughput subject to a latency Service-Level Objective (SLO). The following section describes our proposed scheduler designed to address it.

\section{Proposed solution}
\label{sec:4_scheduler}
\begin{table*}[t!]
\vspace{0.4cm}
    \captionsetup{font=small,labelfont=bf}
    \caption{\normalsize Evaluated DNN models}
    \centering
    \resizebox{\textwidth}{!}{
    \setlength{\tabcolsep}{2pt}
    \begin{threeparttable}
        \begin{tabular}{lllcllrr}
            \toprule
            \textbf{Model} & \textbf{Location} & \textbf{Device} & \textbf{Clock Rate} & \textbf{Accuracy} & \textbf{Latency} & \textbf{FLOPs} & \textbf{\#Params}\\
            \midrule
            MobileNetV2 \cite{sandler2018mobilenetv2}& Low-end & Sony Xperia C5 &  1.69 GHz & 71.85\% & 31 ms & 0.6 B & 3.5 M \\
            EfficientNetLite0 \cite{tan2019effnet}& Mid-tier & Samsung A71 & 2.20 GHz & 75.02\% & 43 ms & 0.8 B & 4.7 M \\
            EfficientNetB0 \cite{tan2019effnet}& High-end & Samsung S20 FE &  2.73 GHz & 77.04\% & 33 ms & 0.8 B & 5.3 M \\
            MobileViT-x-small \cite{mobilevit2022iclr}& High-end & Google Pixel 7 & 2.85 GHz & 74.64\% & 57 ms & 1.1 B & 2.3 M \\
            InceptionV3 \cite{inception2016cvpr}& Server & Tesla T4 GPU & 585 MHz & 78.29\% & 15 ms & 11.4 B & 23.8 M \\
            EfficientNetB3 \cite{tan2019effnet}& Server & Tesla T4 GPU & 585 MHz & 81.49\% & 25 ms & 3.7 B & 12.2 M \\
            DeiT-Base-Distilled \cite{deit2021icml}& Server & Tesla T4 GPU & 585 MHz & 83.41\% & 14 ms & 7.7 B & 86.0 M \\
            \bottomrule
        \end{tabular}
        \begin{tablenotes}
            \small
            \item * See Table 1 in~\cite{oodin2021smartcomp} for the detailed resource characteristics of the target mobile phones.
        \end{tablenotes}
    \end{threeparttable}
    }
    \label{specTable}

\vspace{-0.4cm}
\end{table*}

To address the challenges related to request accumulation and efficient server resource utilization, we introduce \tool, a multi-tenancy-aware scheduler that dynamically adapts the arrival rate of samples from the assisted devices. Based on our previous work, MultiTASC, we kept the component of reconfigurable forwarding decision functions to control the arrival rate and introduced four new techniques that completely change the way the thresholds are updated: \textit{i)}~SLO satisfaction rate updates, \textit{ii)}~continuous threshold reconfiguration, \textit{iii)}~threshold scaling, and \textit{iv)}~server model switching. \tool's internal architecture is presented in Fig.~\ref{fig:scheduler} where all new techniques are visible. 

\subsection{Reconfigurable forwarding decision function}
\label{sec:fwd_dec_func}

Significant research endeavors have been dedicated to assessing the prediction confidence of DNNs, resulting in the development of various approaches~\cite{wang2017idk,guo2017calibration}. In this work, we employ the Best-versus-Second-Best (BvSB) metric~\cite{bvsb2009cvpr}, which quantifies prediction confidence by computing the difference between the top two values in the softmax output of the model (referred to as $P_1$ and $P_2$), as shown in Eq.~(\ref{eq:bvsb}). These values correspond to the highest and second-highest classes that the classifier predicted.
\begin{equation}
    \text{BvSB}\Bigr|_{f(x)} = P_1 - P_2
    \label{eq:bvsb}
\end{equation}
Other metrics, such as top-1 softmax or entropy can be implemented in the system with minimal modifications, potentially leading to different latency-accuracy trade-offs. 

In contrast to the predominant approach employed by most existing cascade systems, which establish fixed decision thresholds during design and maintain them upon deployment, our work introduces an alternative approach. We adopt a \textit{dynamic reconfiguration} scheme for the decision function to cater to the adaptability requirements of our target system. The decision function $d^i(\cdot)$ is defined as shown in Eq.~(\ref{eq:decisionfunc}), with adjustments facilitated by a dynamic scheduler that fine-tunes its parameters at runtime.

\begin{equation}
d^i(f^i_l(x)) =  
 \begin{cases} 
      0 &\text{if} \quad \text{BvSB}\Bigr|_{f^i_l(x)} \geq c_{i,t}\\
      1 &\text{if} \quad \text{BvSB}\Bigr|_{f^i_l(x)} < c_{i,t} 
\end{cases}
\label{eq:decisionfunc}
\end{equation}
where $c_{i,t}$ is the decision threshold of device $i$ at time $t$.
The per-device decision thresholds are exposed to our server-residing scheduler, which adapts them at runtime.

\subsection{SLO satisfaction rate updates} \label{subsec:sloupdates}

Given a latency target, we introduce the SLO satisfaction rate metric as the percentage of samples successfully processed within the designated latency constraint. Latency, in this context, is measured from the initiation of inference on the device until the final result is obtained, either by the device-hosted model or the server-side model in the cases where the sample is forwarded to the server. The SLO satisfaction rate relies significantly on the timely processing of samples forwarded to the server, a factor influenced by the volume of samples that the server must handle.

The SLO satisfaction rate is the main metric we use to measure the smooth operation of the system since a high satisfaction rate can be a good indicator of a responsive queue without congestion. By having a metric that helps us understand the state of the system during runtime, we can tune the influx of samples with high precision by adjusting the reconfigurable forwarding decision functions of each device individually.

Each device calculates the average SLO satisfaction rate during its inference in time windows of $T$ seconds. At the end of every time window, the satisfaction rate for that window is forwarded to the server where the appropriate reconfiguration of the decision function is calculated. By constantly informing the server about the satisfaction rate, we achieve a beneficial trade-off of accuracy in order to maintain the desired SLO satisfaction rate. 

\subsection{Continuous threshold reconfiguration} \label{subsec:threshupdate}

Every $T$ seconds, the device sends an SLO satisfaction rate update to the scheduler. Assuming there is a target SLO satisfaction rate value depending on the user's needs, if the update value is below that optimal value, the server chooses to reduce the number of samples coming from that device. Otherwise, if the update value is greater than the designated optimal value, the server chooses to increase the influx of samples from that device to achieve a higher accuracy. 

\tool leverages the information available through the SLO satisfaction rate updates to view the forwarding decision thresholds as continuous variables. This allows for incredible precision in its ability to identify the optimal threshold for the situation and pair of models, as well as rapid adaptation when needed. The completely revamped update rule is presented in Eq.~(\ref{eq:threshupdate}) and is in stark contrast to MultiTASC's \cite{multitasc2023iscc} update rule where the scheduler had to guess the optimal influx of samples and then use a very slow and imprecise step based approach to converge to it. 
\begin{equation}
\Delta thresh = -a \cdot (SR_{\text{target}} - SR_{\text{update}})
\label{eq:threshupdate}
\end{equation}
where $\Delta thresh$ is the amount by which the threshold will be adjusted, $SR_{\text{target}}$ is the target SLO satisfaction rate for that device, $SR_{\text{update}}$ is the SLO satisfaction rate sent by the device and $a$ is a scaling factor.

\subsection{Threshold scaling} \label{subsec:threshscaling}

The update rule presented in Eq.~(\ref{eq:threshupdate}) demonstrates robust performance in cases where the optimal threshold is either lower or close to the initial threshold of the device.  However, its responsiveness diminishes when facing scenarios characterized by substantial underutilization of server resources, necessitating a rapid threshold increase. To accommodate such cases, we introduced a scaling component to the threshold update. Alg.~\ref{multalg} presents the proposed approach. Following the threshold update from Eq.~(\ref{eq:threshupdate}), the updated threshold is subsequently either scaled by a multiplier $m$ (line~2) if the threshold was increased with the update (\textit{i.e.}~$SR_{\text{update}}>SR_{\text{target}}$), or stays the same otherwise (line~5). In the first case, $m$ is then updated by the rule shown in line~3 while in the second case it is reset to 1. Since for a large number of devices the multiplier is not required, we further incorporate a penalty term (line~3) that considers the number of devices currently active in the system, denoted by $n$.
%
            
%
\begin{algorithm}
    \setcounter{AlgoLine}{0}
    \scriptsize
    \SetAlgoLined
    \LinesNumbered
    \DontPrintSemicolon
    \caption{Multiplier Implementation}
    \label{multalg}
        \KwIn{The targeted satisfaction rate $SR_{\text{target}}$}
        \nonl
        \myinput{The device's satisfaction rate $SR_{\text{update}}$}
        \nonl
        \myinput{The updated threshold $thresh_{\text{updated}}$}
        \nonl
        \myinput{Previous multiplier $m$}
        \nonl
        \myinput{Number of active devices $n$}
        \nonl
        \textbf{Output: } Threshold sent to device $thresh_{\text{final}}$

            
        \uIf{\textup{$SR_{\text{target}} < SR_{\text{update}}$}}{
            $thresh_{\text{final}} = m\cdot thresh_{\text{updated}}$
            
            $m = m\cdot(1 + \frac{0.1}{n})$ 
        }
        \Else{
            $thresh_{\text{final}} = thresh_{\text{updated}}$
            
            $m = 1$
        }
\end{algorithm}
\subsection{Server model switching}
To further improve the adaptability of \tool, we introduce the server model switching feature. Depending on the dynamic conditions, the scheduler can decide to switch to a different server-hosted model whose computational cost-accuracy trade-off better suits the current state. For instance, if a heavy model is being used and the optimal SLO satisfaction rate can't be maintained without significantly dropping the accuracy, \tool opts to switch to a faster model to allow for a greater influx of samples. On the opposite, if a faster model is being used but the system is low load, resulting in underutilization of the server, \tool chooses to switch to a heavier model to achieve a higher accuracy without sacrificing the SLO satisfaction.

The scheduler makes a decision by examining the current thresholds of the devices. To switch from a heavy model to a faster one, every device's threshold in a single tier must be below a certain value. On the other hand, to switch from the fast model to the heavier one, every device's threshold must be above a certain value depending on its tier. 

Given the set of all device thresholds $\mathcal{C}$ consisting of elements $c^{k}_i \in [0,1]$, where $k$ is the device tier (\textit{i.e.} $k \in \mathcal{K} = \{low,mid,high\}$) and $i \in \mathcal{D}^{k}$ corresponds to the i-th device in this tier, we can define the model switch decision as follows: for $S(\mathcal{C})=-1$ a switch to a faster model is needed, for $S(\mathcal{C})=+1$ a switch to a heavier model is needed and for $S(\mathcal{C})=0$ the current model is the optimal. Formally:
\begin{equation*}
S(\mathcal{C}) =  
 \begin{cases} 
      -1 &\text{if} \quad \exists ~k \in \mathcal{K}: ~c^{k}_i < c_{\text{lower}}, ~\forall ~i\in \mathcal{D}^k \\
      +1 &\text{if} \quad c^{k}_i > c_{\text{upper}}^{k}, ~\forall k \in \mathcal{K}, ~\forall ~i\in \mathcal{D}^k \\
      ~0, &\text{otherwise} 
\end{cases}
\label{eq:modelswitch}
\end{equation*}
Upper and lower limits $c_{\text{upper}}^{k}$ and $c_{\text{lower}}$ are set after a thorough examination of cascade results on a training set.

\section{Evaluation}
\label{sec:5_evaluation}
\subsection{Experimental setup}
To evaluate the performance of \tool, we built a prototype in Python 3.9 utilizing TensorFlow 2.9.1. The edge server component  hosts an NVIDIA Tesla T4 GPU, Intel(R) Xeon(R) 2.30GHz CPU and
12GB of RAM, allowing it to run heavy, state-of-the-art models with high accuracy within the evaluated latency targets. On the device components' side, we used mobile devices spanning three different tiers, namely high, mid, and low-end, using Samsung S20 FE, Samsung A71 and Sony Xperia C5 Ultra, respectively. We also used the Google Pixel 7 as a high-end device for the purpose of evaluating transformer models. For on-device execution, we used TensorFlow Lite and targeted the CPU of the respective mobile device, as it is still the most widely-used approach~\cite{smart2021imc, exploring2023iscc}.
\begin{figure*}[t!]
\includegraphics[width=\textwidth]{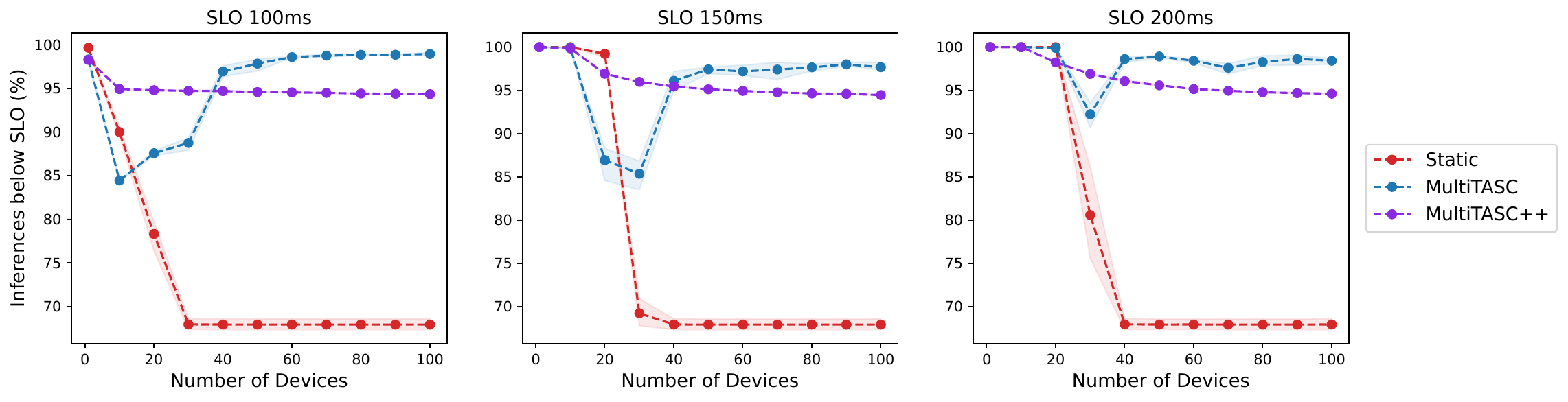}
\caption{SLO satisfaction rate for InceptionV3 - MobileNetV2.}
\label{plot:IncMobSLO}
\end{figure*}
\begin{figure*}[t!]
\includegraphics[width=\textwidth]{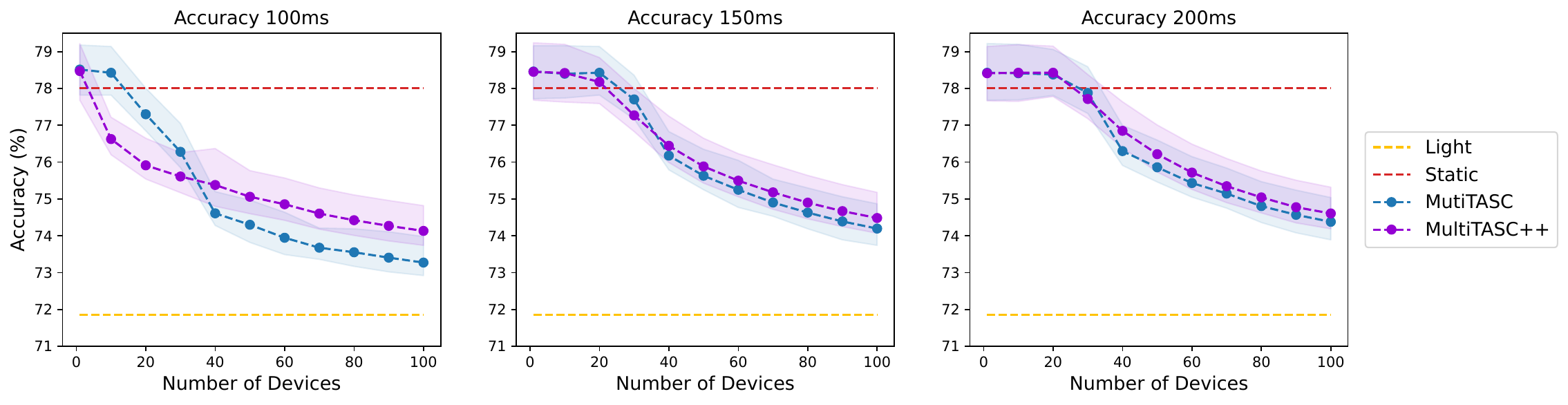}
\caption{Accuracy for InceptionV3 - MobileNetV2.}
\label{plot:IncMobAcc}
\end{figure*}
\begin{figure*}[t!]
\includegraphics[width=\textwidth]{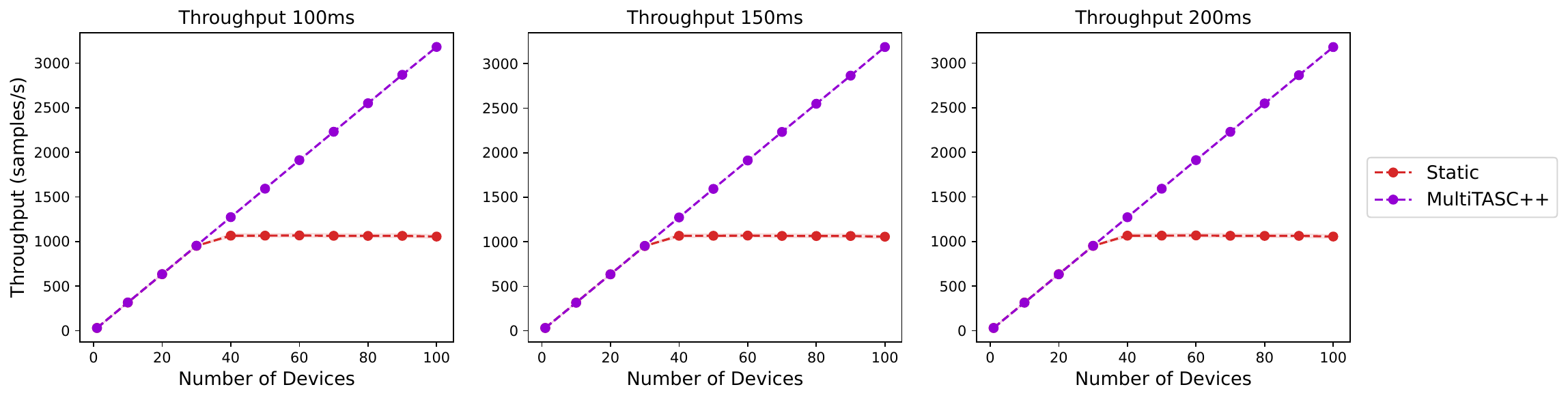}
\caption{Throughput for InceptionV3 - MobileNetV2}
\label{plot:IncMobTP}
\end{figure*}
Different models were chosen for each tier, with the aim of having the highest accuracy possible while taking into account the device's computational resources. We measured the average inference latency of each model on the respective device across 200 runs with a batch size of 1. We followed the same process to measure the average server inference latency across different batch sizes and used this data to conduct simulation-based experiments. The experiments targeted a variety of scenarios and mainly focused on the SLO latency target and accuracy metrics. Communication between the devices and the edge server component was established using the AMPQ protocol, following the widely-used practice for communication between IoT devices. The protocol was implemented through the AMPQStorm library which allows for thread-safe execution. 

To fully take advantage of the server's computational resources and boost throughput, it is important to use batching, \textit{i.e.}~processing multiple samples at the same time. To avoid the latency that would arise from waiting for the request queue to reach a specific batch size, we employ dynamic batching~\cite{ali2020batch}. With dynamic batching, we use the maximum batch size that is feasible with the current request queue length. Available batch sizes are $\mathcal{B}$$=$$\{1,2,4,8,16,32,64\}$. Due to diminishing returns, in some cases we use a lower maximum batch size, \textit{e.g.}~with EfficientNetB3 a batch size of 16 provides a higher throughput and lower latency than a batch size of 32 and above.

\textbf{Models and datasets:} 
In our experiments, we target the task of image classification with 1k classes. Concretely, we used the 50k-images validation set of the ImageNet dataset~\cite{fei2009imagenet}. Table \ref{specTable} presents the evaluated models, as well as their accuracy and inference latency on the different execution points. We obtained the CNN models from TensorFlow Hub while the transformer models from Hugging Face. All models are pretrained on ImageNet's training set. We deploy MobileNetV2, EfficientNetLite0 and EfficientNetB0 to the low, mid and high-end client devices respectively. On the server side, we use InceptionV3 and EfficientNetB3 to explore the differences between a lower-accuracy, higher-throughput model and a higher-accuracy, lower-throughput model, respectively. 

We also evaluated our scheduler using transformer models both as device and server-hosted models. We deployed MobileVit-x-small on Google Pixel 7 and DeiT-Base-Distilled on the server. As the transformer architecture has only recently started gaining traction, available models are not as efficient as their CNN counterparts on most mobile devices. As such, we utilize a computationally-powerful flagship phone, like Pixel 7, to achieve acceptable latency. Nevertheless, as transformer models become more efficient~\cite{edgevits2022eccv}, they are expected to soon become broadly deployed across mobile devices.

\textbf{Evaluation settings:} To assess our system across deployment setups, we focused on two distinct cases: \textit{i)}~a homogeneous scenario, which comprises devices of equal processing capabilities that host the same local model, and \textit{ii)}~a heterogeneous scenario, which comprises devices of diverse processing capabilities, with each device hosting a model tailored to its respective tier. We also conduct experiments as part of the homogeneous scenario to evaluate the transformer models, the server model switching technique, as well as a scenario that emulates a realistic setting of intermittent device participation, \textit{i.e.}~where devices drop in and out during inference. 

\begin{figure*}[t]
\includegraphics[width=\textwidth]{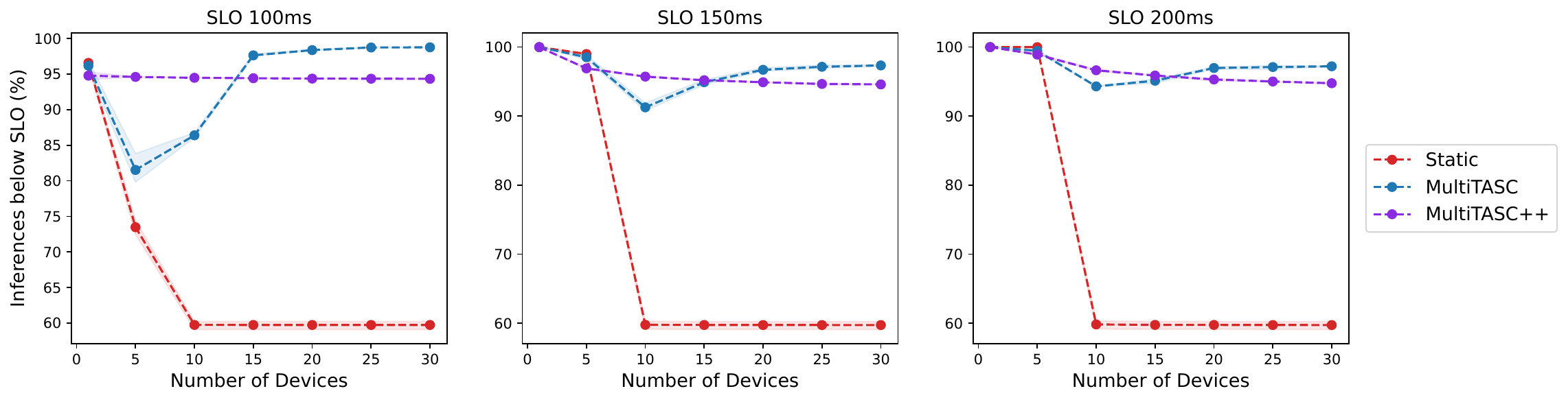}
\caption{SLO satisfaction rate for EfficientNetB3 - MobileNetV2.}
\label{plot:EffMobSLO}
\end{figure*}

\begin{figure*}[t]
\includegraphics[width=\textwidth]{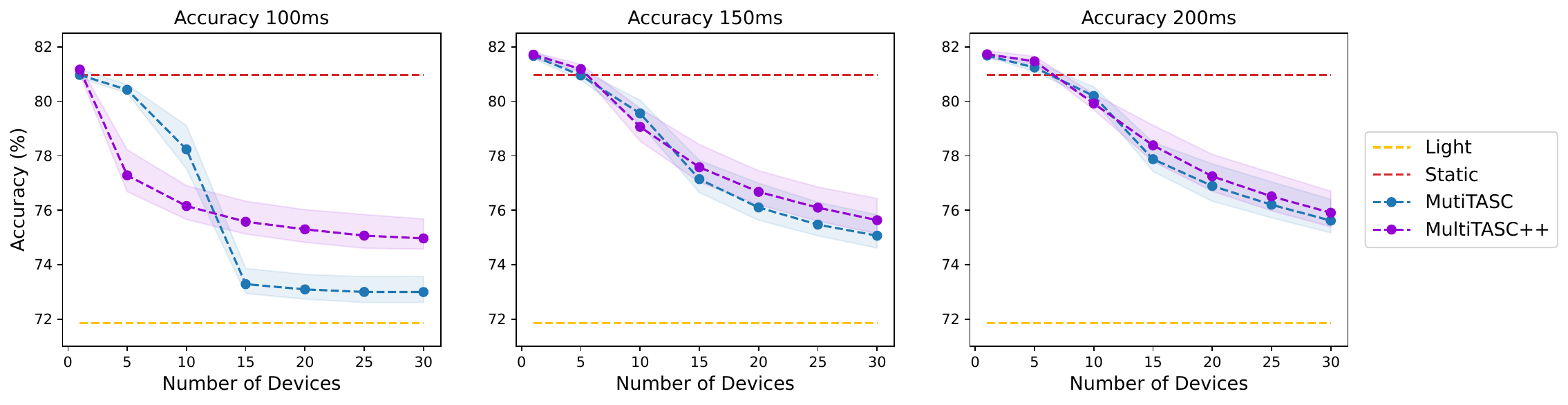}
\caption{Accuracy for EfficientNetB3 - MobileNetV2.}
\label{plot:EffMobAcc}
\end{figure*}

\begin{figure*}[t!]
\includegraphics[width=\textwidth]{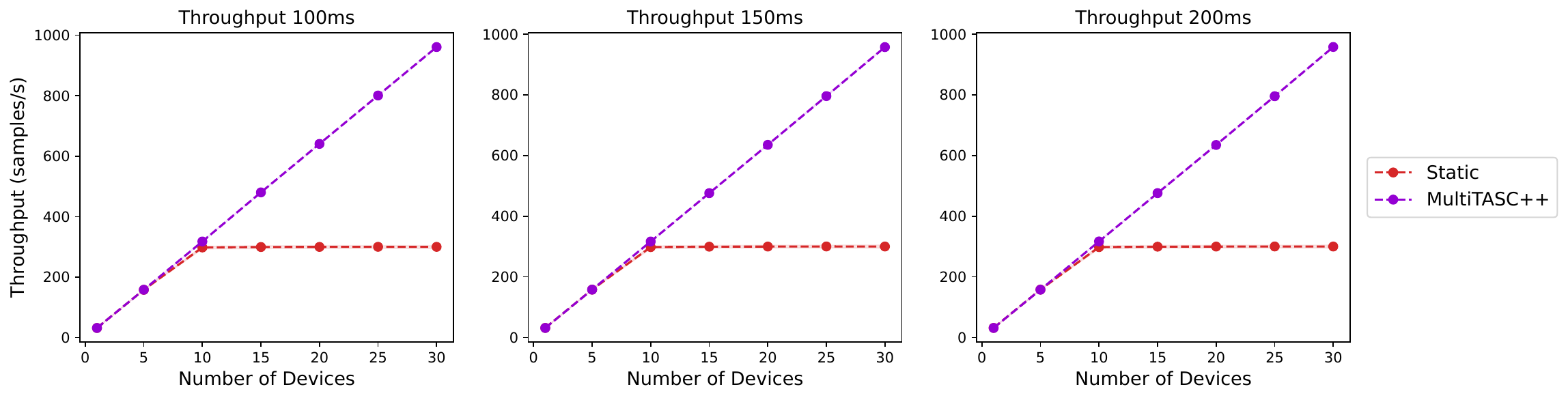}
\caption{Throughput for EfficientNetB3 - MobileNetV2.}
\label{plot:EffMobTP}
\end{figure*}
For general evaluation in the homogeneous scenario, all devices are of the same tier, using Sony Xperia C5 Ultra with MobileNetV2. We selected this tier, as it represents the configuration with the minimum latency, thereby imposing the greatest challenge on the scheduler. Additionally, it accentuates the substantial disparity in accuracy between the device-hosted model and the full cascade accuracy, effectively showcasing the cascade architecture's potential. 
In the heterogeneous scenario, all three tiers of devices were deployed in equal percentage. Finally, to assess the scheduler's performance when using transformer models, we used a separate tier of Pixel 7 devices running MobileViT-x-small. 

\begin{figure*}[t!]
\includegraphics[width=0.75\textwidth]{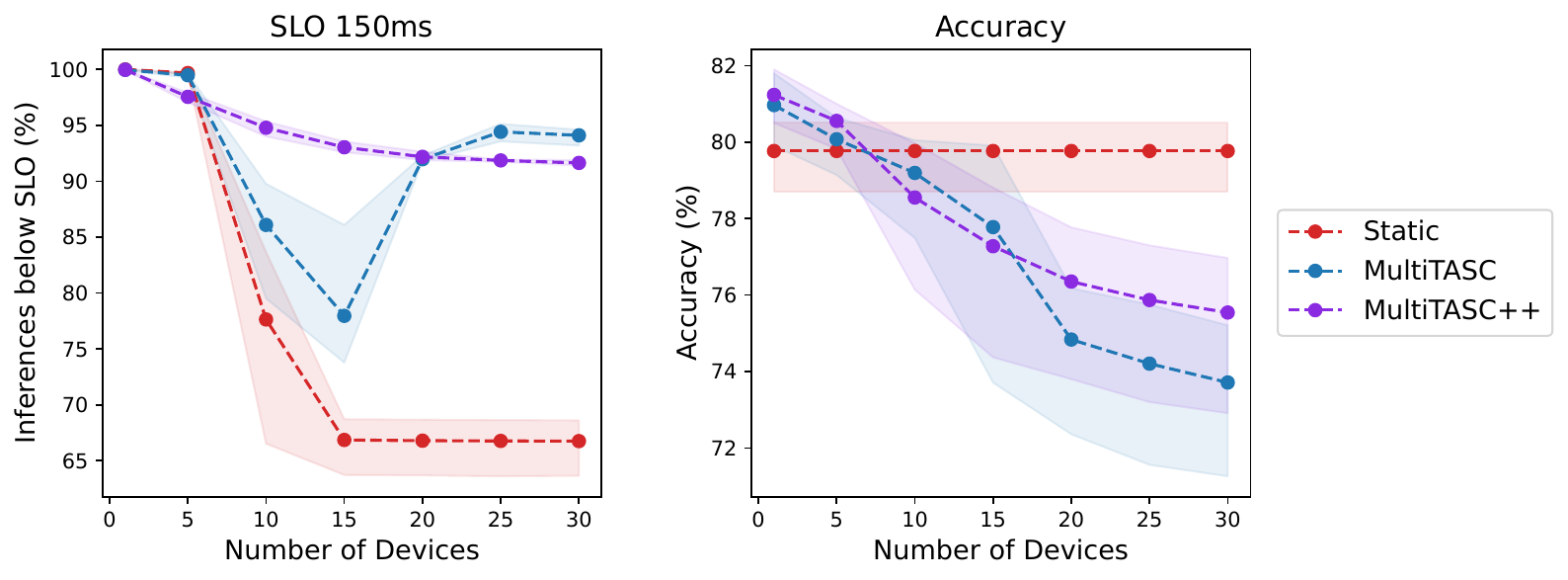}
\centering
\caption{EfficientNetB3 - MobileNetV2 with 1000 samples.}
\label{plot:EffMob1000}
\end{figure*}

In all scenarios but one, the dataset of each device consisted of 5000 randomly selected samples from the last 40000 images of ImageNet’s validation set. The exception consisted of a distinct scenario where the dataset consisted of only 1000 samples. We followed this approach in order to more clearly highlight a limitation of MultiTASC~\cite{multitasc2023iscc} where the SLO satisfaction rate was below its required value due to slow convergence of the threshold reconfiguration. As shown later in Fig.~\ref{plot:EffMob1000}, \tool alleviates effectively this issue. 

Three different seeds were used to sample the data and the average values of each metric, alongside their minimum and maximum, are reported. The metrics used for the evaluation are: the system throughput capturing the system's processing rate in samples/s, the average accuracy across devices, the latency SLO satisfaction rate for 100, 150 and 200 ms SLOs, and the scalability in terms of number of devices.

\textbf{Baselines:} As a baseline, we use Static, a scheduler with statically selected thresholds that remain fixed during runtime. To tune the static threshold, we use the first 10000 images of ImageNet’s validation 
set as our calibration set and evaluate all cascade model pairs in terms of accuracy and forwarding probability. As such, we tune the threshold so that approximately 30\% of samples are forwarded to the heavy model, providing a balanced accuracy-latency trade-off. In cases where that threshold yielded an accuracy loss of more than 1~pp compared to the highest achievable cascade accuracy, we used the lowest threshold that satisfied the 1~pp limit. This baseline is equivalent to a set of state-of-the-art cascades~\cite{li2021appealnet,wang2017idk,kouris2018cascade}. We also compare with \mbox{MultiTASC}~\cite{multitasc2023iscc}, which constitutes the current state-of-the-art scheduling method for multi-device cascades.
\subsection{Evaluation of performance}
In this section, we assess the performance of our scheduler compared to Static and MultiTASC across all scenarios. We set the target SLO value to 95, meaning that we aim for 95\% of samples to finish inference within the latency target, independent of whether they are forwarded to the server or stay on the device. We do not aim for 100\% so that the system has some leeway to trade SLO satisfaction rate percentage for accuracy. Here, we highlight a significant improvement introduced by \tool. Our new scheduler allows us to set and consistently maintain a target satisfaction rate regardless of the prevailing conditions, addressing a limitation present in MultiTASC.

We also set the time window $T$ to 1.5~s and the scaling variable $a$ of the continuous threshold to $0.005$. Compared to MultiTASC, the amount of variables to initialize and the computational effort required to determine their values have notably decreased. This streamlines the scheduler's deployment process, while delivering substantially improved and consistent results. Furthermore, the new continuous threshold update scheme allows for the SLO targets to be chosen independently for each device, contrasting MultiTASC where all devices had to agree on the same latency target during the initialization of the scheduler.
\begin{figure*}[t]
\includegraphics[width=\textwidth]{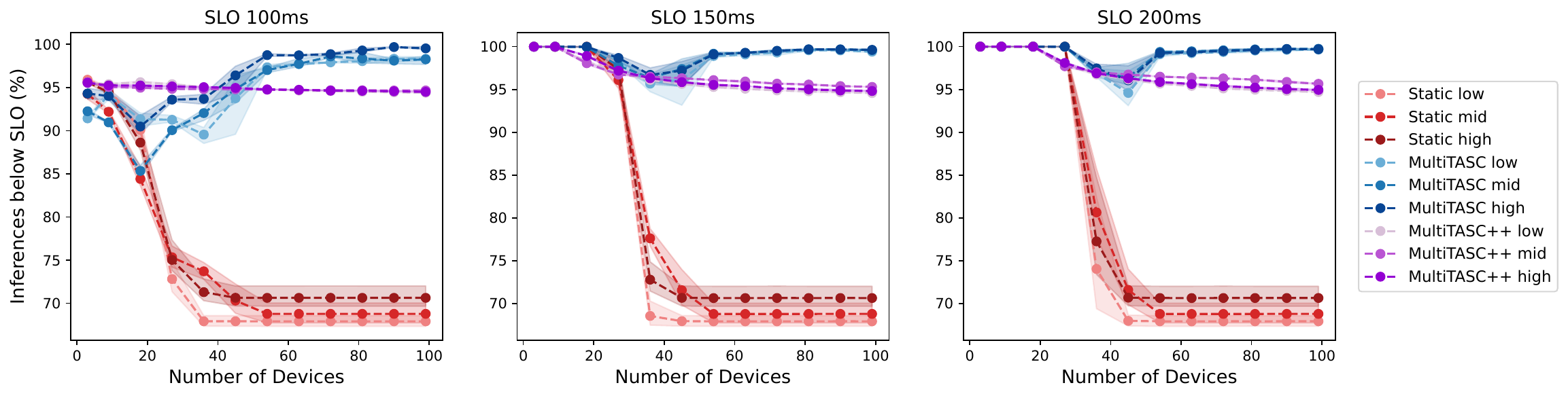}
\caption{SLO satisfaction rate for InceptionV3 - different devices.}
\label{plot:IncDiffSLO}
\end{figure*}
\begin{figure*}[t!]
\includegraphics[width=\textwidth]{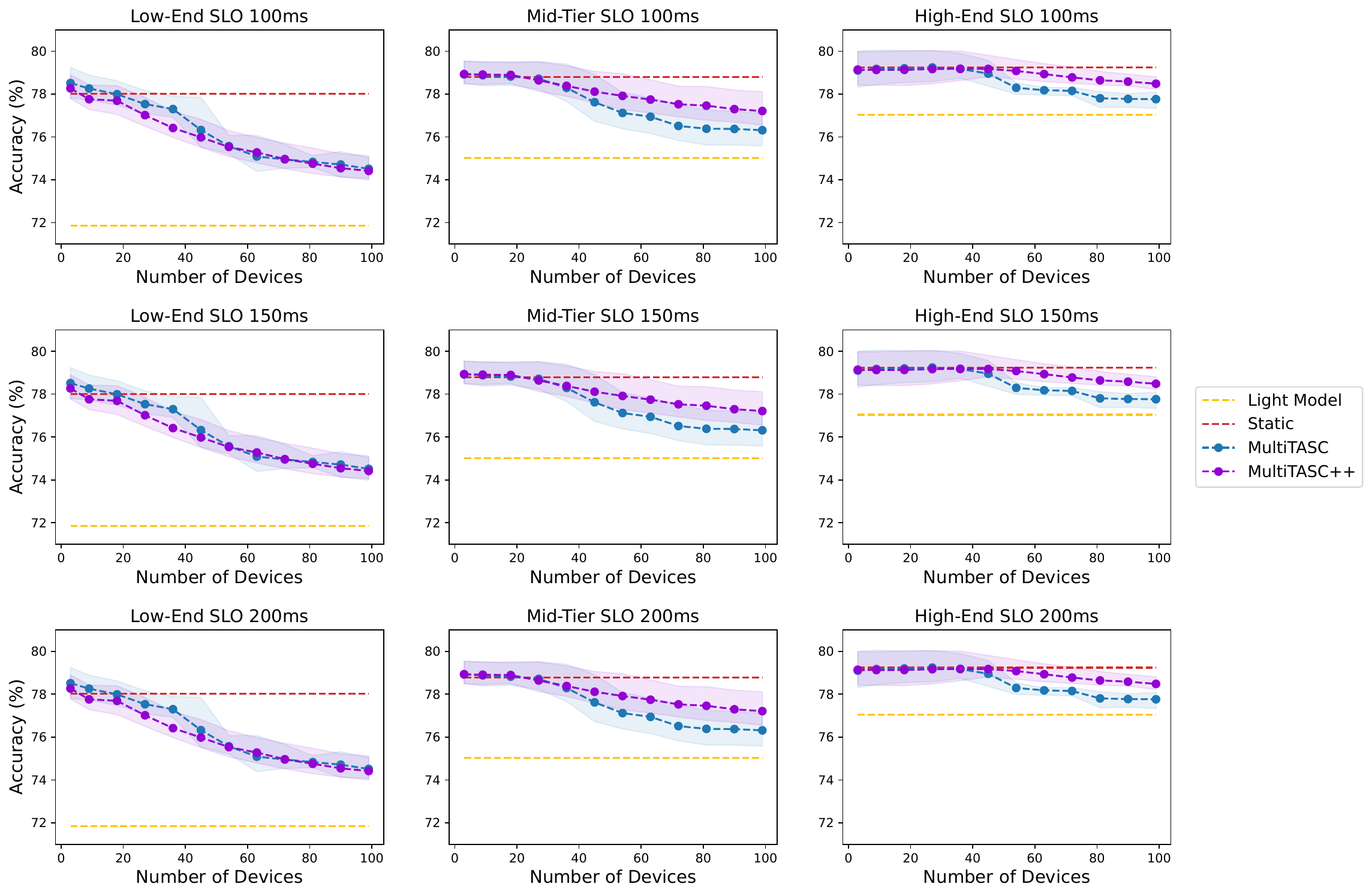}
\caption{Accuracy for InceptionV3 - different devices.}
\label{plot:IncDiffAcc}
\end{figure*}
\subsubsection*{A. Homogeneous scenario}
Fig.~\ref{plot:IncMobSLO} shows the SLO satisfaction rate as the number of devices increases with InceptionV3 hosted on the server. The figure includes the
\tool, Static and MultiTASC approaches. Notably, \tool manages to consistently maintain the satisfaction rate close to or above 95\%. In contrast, Static experiences rapid degradation, with the majority of the results of the forwarded samples not returning within the latency target for 25-40 devices and above. On the other hand, MultiTASC exhibits a dip in the range of approximately 5 to 40 devices, followed by an overcorrection that achieves a 100\% satisfaction rate resulting in a needless degradation in accuracy. This dip occurs due to MultiTASC opting to use batch size as a metric for congestion, which proves to be suboptimal. Pairing that with a static step update rule, the scheduler was neither able to accurately predict the state of the system nor adapt with the speed that is required in borderline system states. This is mitigated in MultiTASC++, where the scheduler is more accurately aware of the state of the system through the SLO updates from the devices (Section~\ref{subsec:sloupdates}), while also being able to adapt more quickly through the techniques of continuous threshold reconfiguration (Section~\ref{subsec:threshupdate}) and threshold scaling (Section~\ref{subsec:threshscaling}).

Fig.~\ref{plot:IncMobAcc} presents the full cascade's accuracy in comparison to the baselines as the number of devices increases.~\tool effectively strikes a better balance between accuracy and the maintenance of the desired 95\% SLO satisfaction rate. In comparison to Static,~\tool achieves a higher accuracy for a smaller number of devices where the server is being underutilized, while for larger amounts of devices, it chooses to trade off accuracy in order to sustain the satisfaction rate. Importantly, even though the accuracy is lower compared to the static approach, it is still substantially higher than the device-hosted model's accuracy, justifying the use of a cascade architecture. Compared to MultiTASC, our scheduler achieves higher accuracy across all cases, with the exception of the instances where MultiTASC's satisfaction rate performance (Fig.~\ref{plot:IncMobSLO}) dips below the desired 95\%.
\begin{figure*}[t]
\includegraphics[width=\textwidth]{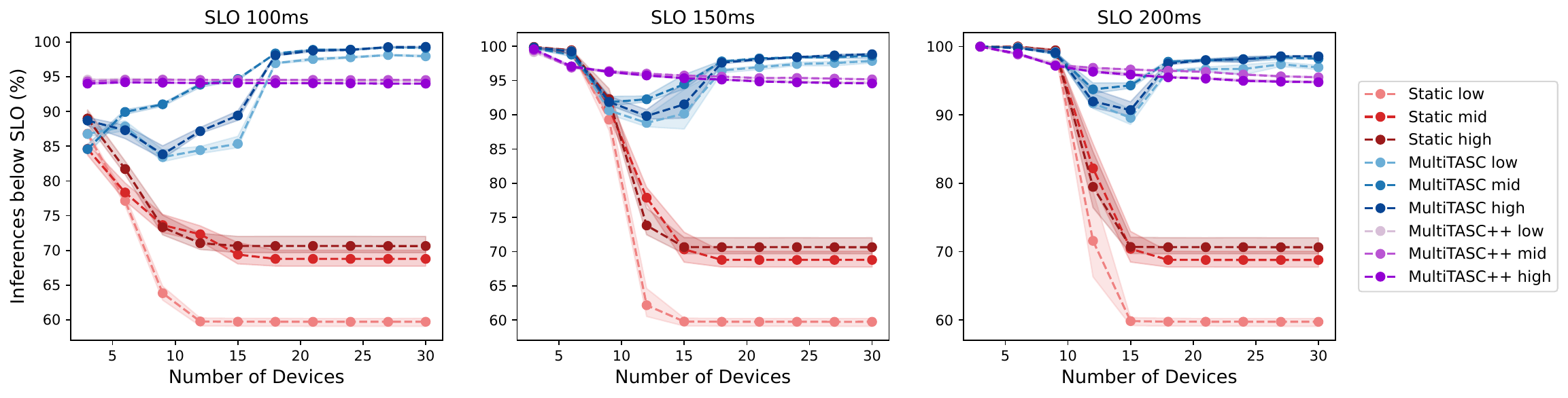}
\caption{SLO satisfaction rate for EfficientNetB3 - different devices.}
\label{plot:EffDiffSLO}
\end{figure*}
\begin{figure*}[t!]
\includegraphics[width=\textwidth]{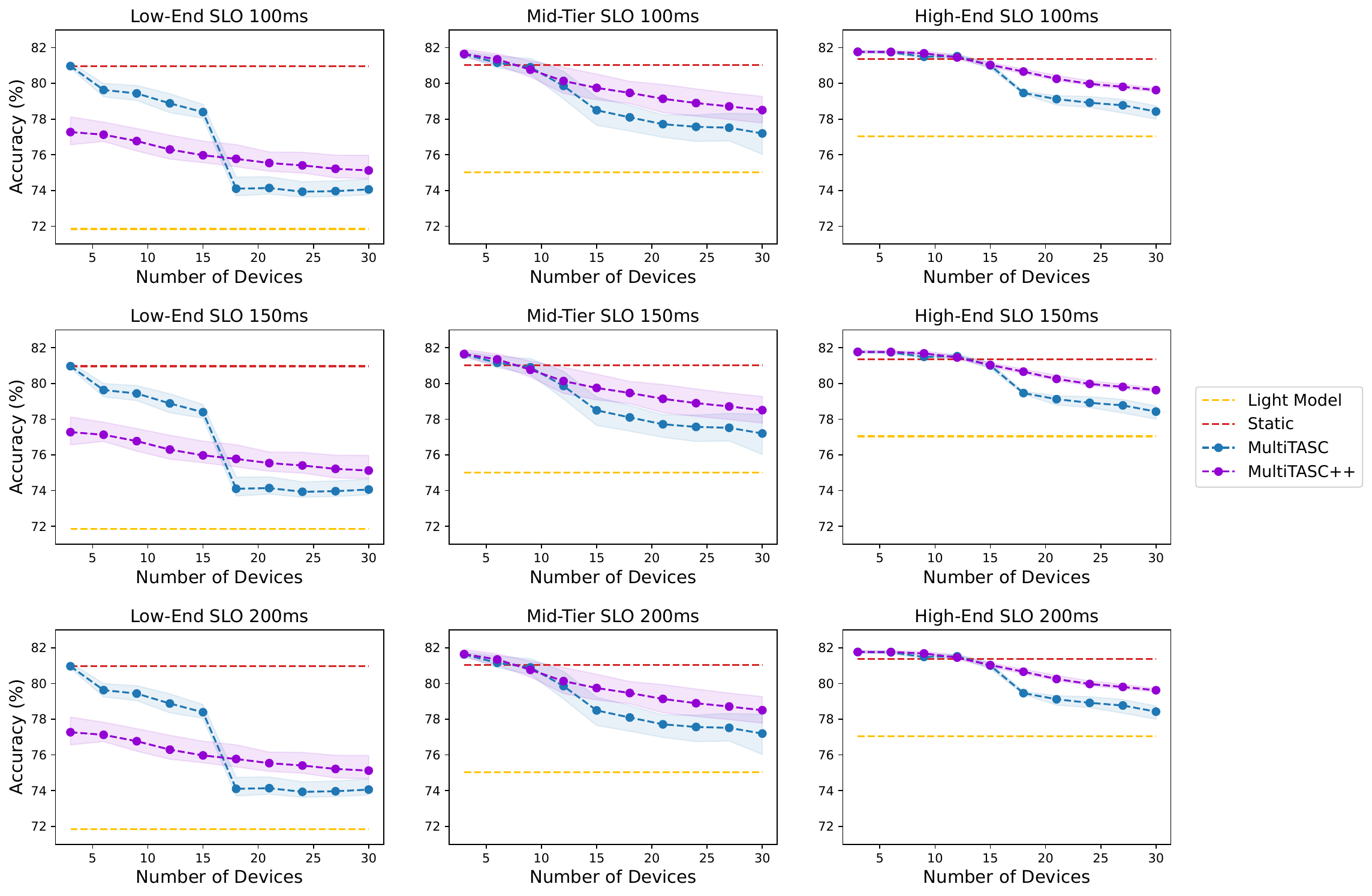}
\caption{Accuracy for EfficientNetB3 - different devices.}
\label{plot:EffDiffAcc}
\end{figure*}
Fig.~\ref{plot:IncMobTP} shows the system throughput difference between \tool and Static as the number of devices increases. This figure shows the need for dynamic runtime adaptation, since, while Static stagnates at 1000 samples per second, \tool manages to keep the linear increase of system throughput as the number of devices rises for all SLO targets.
\begin{figure*}[t!]
\includegraphics[width=\textwidth]{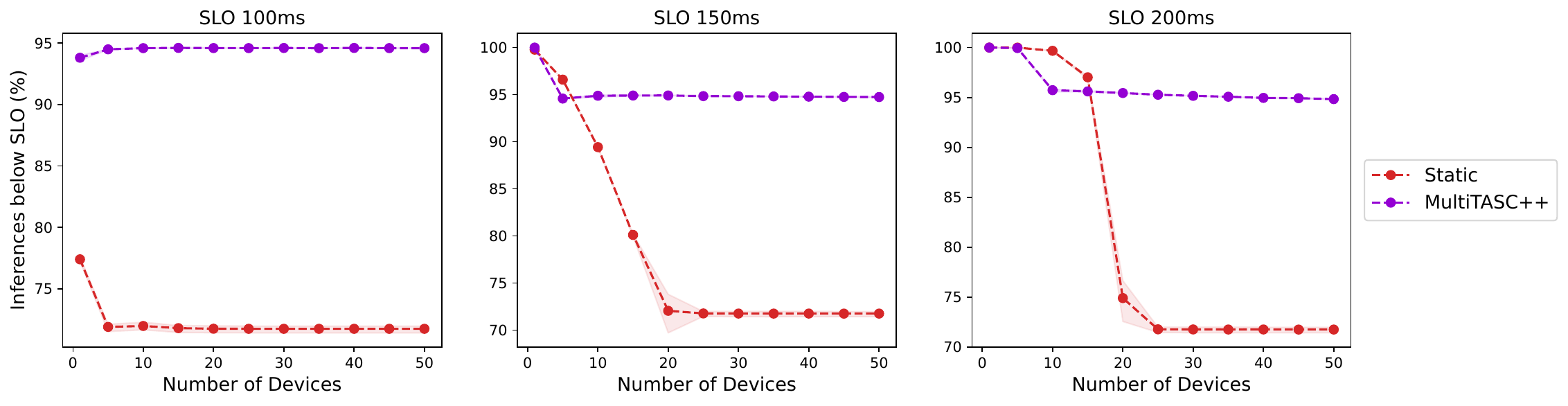}
\caption{SLO satisfaction rate for DeiT-Base-Distilled - MobileViT-x-small.}
\label{plot:ViTDeiTSLO}
\end{figure*}
\begin{figure*}[t!]
\includegraphics[width=\textwidth]{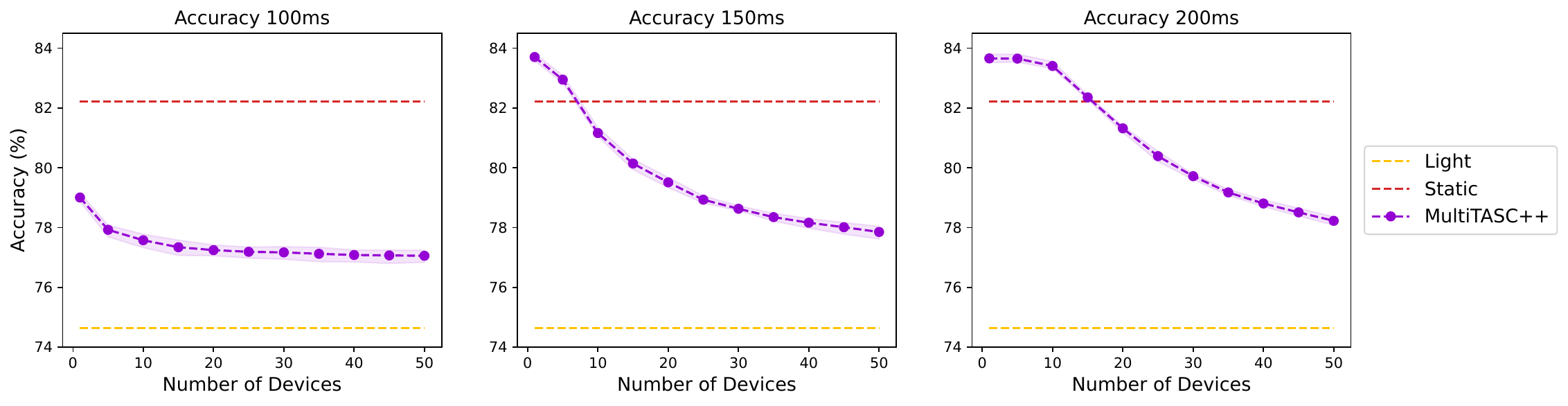}
\caption{Accuracy for DeiT-Base-Distilled - MobileViT-x-small.}
\label{plot:ViTDeiTAcc}
\end{figure*}
Fig.~\ref{plot:EffMobSLO} and Fig.~\ref{plot:EffMobAcc} depict the achieved SLO satisfaction rate and accuracy, when EfficientNetB3 is deployed on the server. A similar dip to the one in Fig.~\ref{plot:IncMobSLO} can be observed for MultiTASC, between approximately 5 and 20 devices. The dips in both figures reach values as low as 80\%, which represents unacceptable delays in 15pp more samples than the target. We can also observe a more significant difference in accuracy, especially when targeting a 100 ms SLO, due to the lower attainable throughput of the heavier EfficientNetB3. Notably, even with a substantial number of devices, the collaborative cascade architecture, when paired with \tool, significantly improves upon the accuracy of the on-device model while preserving consistent responsiveness. In contrast to MultiTASC, besides optimizing the trade-off between satisfaction rate and accuracy, our approach also minimizes the variance between different seed runs, a crucial element in ensuring a robust high-quality service. Fig.~\ref{plot:EffMobTP} shows similar results to Fig.~\ref{plot:IncMobTP}, where Static converges to a system throughput of around 300 samples per second while \tool manages to maintain a linear increase of system throughput as the number of devices increases.

Fig.~\ref{plot:EffMob1000} shows the SLO satisfaction rate and accuracy trends with an increasing number of devices, utilizing a reduced dataset of 1000 samples, as opposed to the previous 5000 and with a lenient 150 ms SLO. A noticeable distinction becomes apparent when comparing the evaluated approaches. MultiTASC converges slowly to a threshold that satisfies the SLO, evident by the results between 10 to 20 devices, where the satisfaction rate reaches a low point of 75\%. In contrast, \tool consistently delivers nearly identical results to those observed in the prior experiment.
\subsubsection*{B. Heterogeneous scenario}
Fig.~\ref{plot:IncDiffSLO} and Fig.~\ref{plot:IncDiffAcc} present a comprehensive comparison of SLO satisfaction rate and accuracy across \tool, MultiTASC, and Static in a heterogeneous device environment. In this setting, InceptionV3 serves as the server-side model. We report the performance metrics separately for each device tier. 
\begin{figure*}[t]
\includegraphics[width=0.8\textwidth]{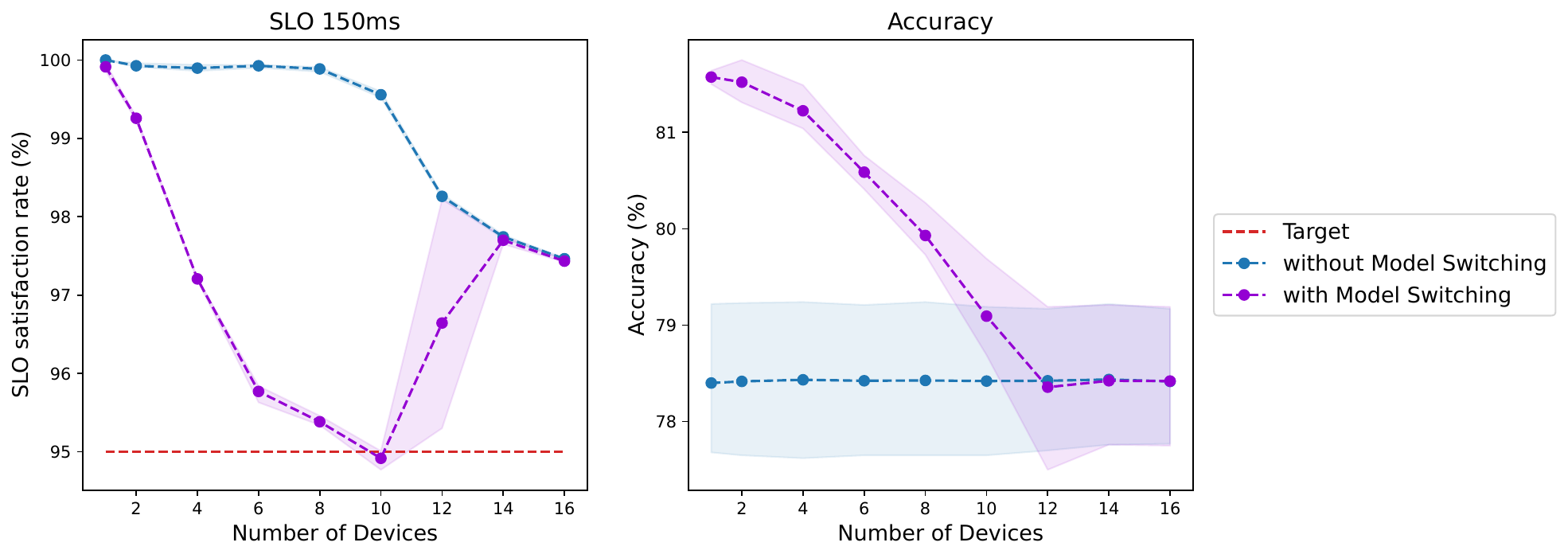}
\centering
\caption{Model switching with InceptionV3 initialization.}
\label{plot:ModelSwitchInc}
\end{figure*}
\begin{figure*}[t]
\includegraphics[width=0.8\textwidth]{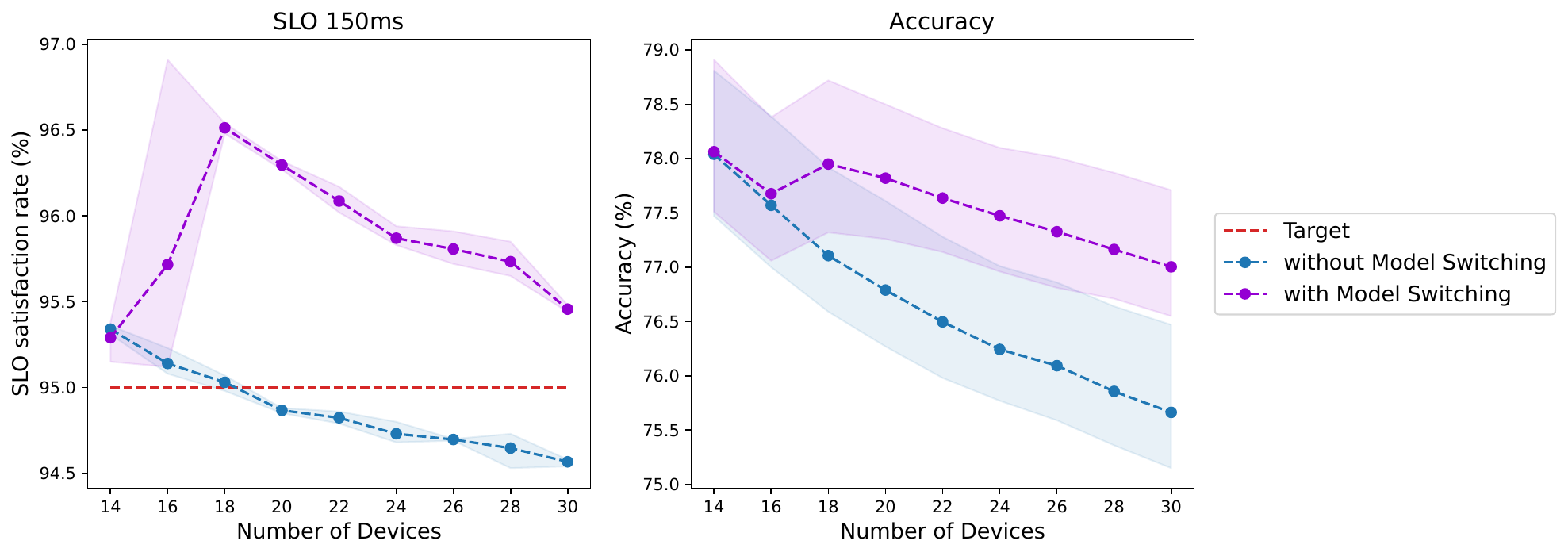}
\centering
\caption{Model switching with EfficientNetB3 initialization.}
\label{plot:ModelSwitchEff}
\end{figure*}
Similar to the homogeneous scenario, our observations underscore the limitations of the static approach, which experiences a critical failure beyond a certain number of devices. Conversely, both \tool and MultiTASC effectively maintain high satisfaction rates. Notably, our proposed scheduler exhibits superior consistency and efficiency in trading accuracy for satisfaction rate, as it robustly maintains satisfaction rates precisely at the targeted level. MultiTASC, while providing operational viability compared to Static for more devices, it also introduces significant variance that can undermine the quality of service. Additionally, it does not effectively utilize the given satisfaction rate allowance, resulting in lower accuracy levels than potentially achievable.

Furthermore, \tool mitigates the dip experienced by MultiTASC within the range of approximately 5 to 40 devices. In particular, when it comes to mid and high-end tier devices, \tool achieves significant accuracy gains over MultiTASC, while accuracy remains relatively consistent for low-end devices between both schedulers.

Fig.~\ref{plot:EffDiffSLO} and Fig.~\ref{plot:EffDiffAcc} present the same comparison with EfficientNetB3 as the server-side model. The results and conclusions obtained are similar to those from the previous experiment, where InceptionV3 was the server model.
\begin{figure*}[t!]
\includegraphics[width=\textwidth]{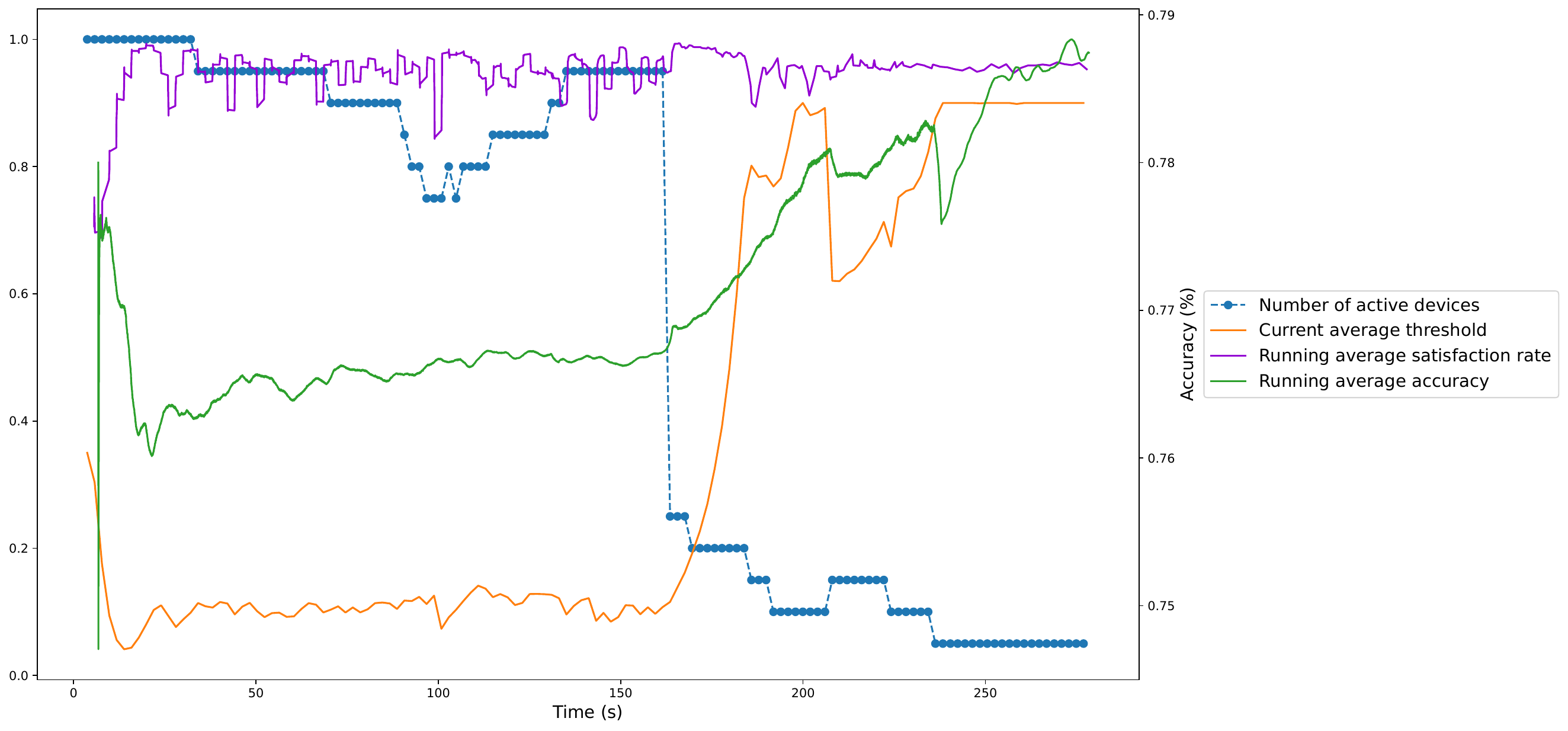}
\caption{\tool's behavior under intermittent device participation with dynamic threshold.}
\label{plot:SimDynamic}
\end{figure*}
However, it is worth emphasizing that the variance and fluctuations observed in MultiTASC's satisfaction rate are even more pronounced in this scenario. In terms of accuracy, MultiTASC achieves a higher accuracy with low-end devices, particularly in the range of 0 to 15 devices, but subsequently lags behind, mirroring the patterns observed in mid and high-end devices. This phenomenon can be attributed to the low-end devices' lower latency, which significantly contributes to congestion, necessitating faster threshold adjustments. It is also worthy to note that \tool achieves a higher accuracy for light devices when it comes to larger numbers of devices.
\subsubsection*{C. Evaluation on transformers}
Fig.~\ref{plot:ViTDeiTSLO} and Fig.~\ref{plot:ViTDeiTAcc} offer an examination of SLO satisfaction rates and accuracy in response to an increasing number of devices using \tool and Static. In this scenario, the device model is the mobile-grade MobileViT-x-small transformer, while the server model is DeiT-Base-Distilled.

The outcomes closely resemble those observed in previous scenarios, showcasing the ability of our scheduler to generalize to transformer architectures, even though its design was not tailored to them. Any apparent disparities in the results compared to the CNN-based experiments can be attributed to differences in inference latencies and model accuracies.
\subsubsection*{D. Evaluation of model switching}
Fig.~\ref{plot:ModelSwitchInc} presents a comparative analysis of the SLO satisfaction rate and accuracy with the server model switching feature enabled and disabled. The scheduler is initialized with the InceptionV3 model, targeting a 150 ms latency. Up to 12 devices, during runtime \tool decides to switch to a heavier model, namely EfficientNetB3. This achieves a substantially higher accuracy while keeping the satisfaction rate above the 95\% target. For 14 devices and above, the switch is no longer efficient and InceptionV3 is used across all samples. We should note here that the model switching feature was not used in previous experiments so that our update rule could be fairly evaluated against MultiTASC without other techniques affecting the performance.
\begin{figure*}[t!]
\includegraphics[width=\textwidth]{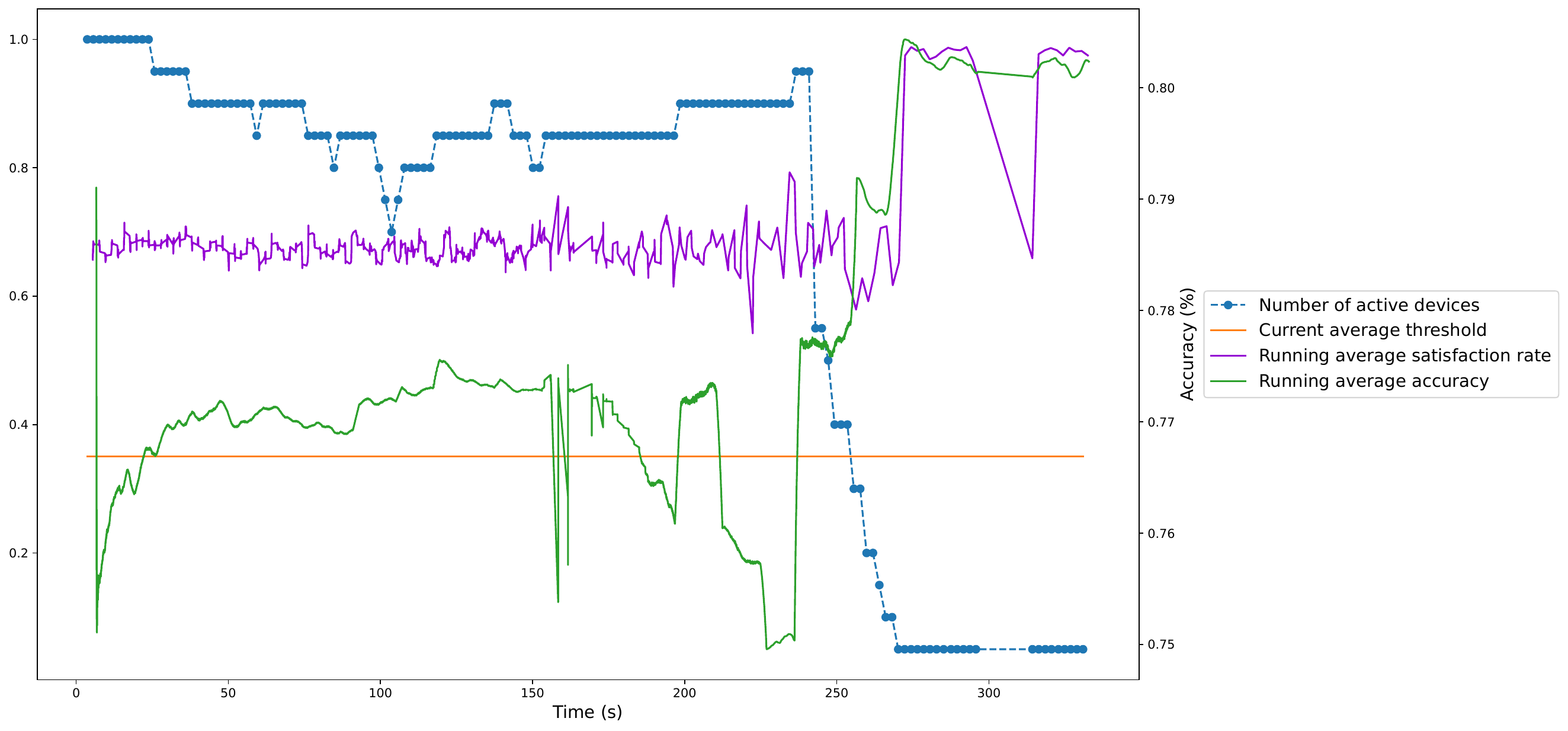}
\caption{\tool's behavior under intermittent device participation with static threshold.}
\label{plot:SimStatic}
\end{figure*}
Fig.~\ref{plot:ModelSwitchEff} shows a similar evaluation for the scheduler, with the server-side model initialized to EfficientNetB3 and targeting a 150 ms latency. Once again, the satisfaction rate consistently exceeds the 95\% target, accompanied by an observable accuracy enhancement when the model switches to InceptionV3. In scenarios featuring 14 and 16 devices, model switching does not occur, leading to outcomes similar to instances without model switching.
\subsubsection*{E. Intermittent device participation}
In this experiment, we emulate a realistic setting where 20 devices run simultaneously, each bearing a 50\% probability of going offline. We target low-tier devices with EfficientNetB3 as the server-hosted model. The point at which a device goes offline follows a normal distribution with a mean $\mu = N/2$ and a standard deviation $\sigma = N/5$ where $N$ is the total number of samples. The duration for which a device remains offline adheres to an alpha distribution with a shape parameter $\alpha = 60$ seconds.

Fig.~\ref{plot:SimDynamic} shows an overview of the dynamics in this experiment. It illustrates the fluctuation in the number of active devices over time, the average threshold maintained across devices, the running SLO satisfaction rate, and the average running accuracy for the currently active devices. We note that the percentage of active devices, the average threshold, and the running satisfaction rate are represented on the left y-axis scale, while the running accuracy is on the right y-axis scale. Several key observations emerge from this visualization.

First, we notice an inverse correlation between the threshold and the number of active devices. Initially, the threshold rapidly decreases to accommodate the need for maintaining the satisfaction rate. Subsequently, as the number of active devices diminishes, the threshold is increased.

Furthermore, a direct correlation is observed between the increase in threshold and the gain in running accuracy. This emphasizes the trade-off mechanism facilitated by the scheduler to optimize system performance.

Notably, the running SLO satisfaction rate consistently maintains a level exceeding approximately 95\% throughout the duration of the experiment, underlining \tool's effectiveness in ensuring the fulfillment of the specified service-level objectives.

Fig.~\ref{plot:SimStatic} shows the results of another experiment with different initializations when it comes to devices, leading to a different plot line representing the number of active devices compared to the previous experiment. In this scenario, however, a static threshold is employed. The results from this experiment unveil several insights.

The utilization of a static threshold, set at 0.35, engenders relatively stable running accuracy, as a completely static running accuracy is not realistic. However, it is accompanied by a notable variability in the running satisfaction rate, consistently falling well below the 95\% target. This observation underlines the critical role played by a dynamically adaptive threshold in ensuring that SLOs are met.

Both the running accuracy and the running satisfaction rate eventually converge at approximately 250 seconds when the devices complete their inference tasks. Meanwhile, the threshold and the number of active devices continue to vary. This divergence arises due to extensive congestion within the request queue, resulting in a delay of approximately 30 seconds between the devices completing their inference and the server's return of all requested results. Since approximately 50\% of the devices do not go offline, we expect the number of active devices to drop in half at around 155 seconds where their inference finishes. While this can be observed in Fig.~\ref{plot:SimDynamic} where the system remains responsive due to \tool, in the case of Fig.~\ref{plot:SimStatic} this does not happen. This is due to the congestion in the request queue and the devices staying connected, waiting for the results to return from the server. 

Overall, this examined scenario further highlights the benefit of dynamic threshold adaptation during runtime, particularly in settings characterized by evolving conditions and varying system demands.

\section{Conclusion}
\label{sec:7_conclusion}
In this paper, we presented \tool, a dynamic scheduler designed to address the challenges of congestion in collaborative DNN inference involving a multitude of IoT devices in indoor intelligent spaces. By introducing the concept of the multi-tenant cascade, we have achieved continuous dynamic adaptation of threshold values, optimizing the trade-off between accuracy and service-level objectives during runtime.

Our experimental evaluation, spanning diverse device environments and server-side models, has demonstrated the efficacy of our dynamic scheduler. Moreover, these experiments comparing \tool to our old implementation, led us to the conclusion that the SLO satisfaction rate updates are critical for precisely adhering to the targets set by each device. Equally essential is the adoption of continuous threshold values, paired with a method designed to counteract slow updates, for the scheduler to achieve its full potential. The harmonic cooperation between all of these components led to a scheduler characterized by its consistent ability to maintain the satisfaction rate that is targeted, as well as allowing for system throughput to scale, while simultaneously enhancing performance across various device tiers and workloads, surpassing the abilities of its predecessor, MultiTASC.

Moreover, the implementation of model switching has proven effective, allowing for the scheduler to adapt to scenarios where a different server-hosted model leads to greater efficiency.

Lastly, by emulating a realistic deployment scenario with intermittent partial participation of devices, we demonstrated our dynamic scheduler's versatility, which can adapt to dynamic device availability and tune the execution configuration in a timely manner. This adaptability is vital for optimizing system performance and ensuring that the service-level objectives are met.

While image classification is a well-explored and commonplace task, it would be of interest for future research to shift attention towards other significant tasks, such as speech recognition. Furthermore, the proposed system could be extended by investigating the new challenges of generative inference tasks, such as image and text generation, and designing novel methods in order to enable their deployment in multi-device cascade setups.

\bibliographystyle{IEEEtran}
{\footnotesize
\bibliography{references}
}

\end{document}